\pdfoutput=1
\documentclass[11pt]{article}

\usepackage{coling}

\usepackage{times}
\usepackage{latexsym}

\usepackage[T1]{fontenc}
\usepackage[utf8]{inputenc}

\usepackage{microtype}

\usepackage{inconsolata}

\usepackage{graphicx}
\usepackage{amssymb}
\usepackage{amsmath}
\usepackage{listings}
\usepackage{placeins}
\usepackage{listings}
\usepackage[most]{tcolorbox}

\newtcolorbox{promptenv}{
center,
width=1.0\linewidth,
halign=left,
colframe=black,
colback=white,
fontupper=\small
}

\newtcolorbox{exampleNER}[1][]{colback=black!10!white, colframe=black!50!white, boxrule=0.5pt, frame hidden, width=1\linewidth, sharp corners=all, fontupper=\small\sffamily, fonttitle=\small, center, before skip=1mm, after skip=2mm, top=0.5mm, bottom=0.5mm, left=0mm} 
\definecolor{colorPER}{HTML}{ED1B23}
\newtcbox{\tagPER}{enhanced, nobeforeafter, tcbox raise base, boxrule=0.4pt, top=0mm, bottom=0mm, right=0mm, left=4mm, arc=1pt, boxsep=2pt, before upper={\vphantom{dlg}}, colframe=colorPER, coltext=colorPER, colback=colorPER!10, overlay={\begin{tcbclipinterior}\fill[colorPER!75] (frame.south west) rectangle node[text=white, font=\sffamily\bfseries\tiny,rotate=90] {PER} ([xshift=4mm]frame.north west);\end{tcbclipinterior}}}
\definecolor{colorO}{HTML}{606060}
\newtcbox{\tagO}{enhanced, nobeforeafter, tcbox raise base, boxrule=0.4pt,top=0mm, bottom=0mm, right=0mm, left=3mm, arc=1pt, boxsep=2pt, before upper={\vphantom{dlg}}, colframe=colorO, coltext=colorO!75!black, colback=colorO!10, overlay={\begin{tcbclipinterior}\fill[colorO!75] (frame.south west) rectangle node[text=white, font=\sffamily\bfseries\tiny] {O} ([xshift=3mm]frame.north west);\end{tcbclipinterior}}}

\newtcolorbox{examplecoref}[1][]{colback=black!10!white, colframe=black!50!white, boxrule=0.5pt, frame hidden, width=1\linewidth, sharp corners=all, fontupper=\small\sffamily, fonttitle=\small, center, before skip=1mm, after skip=2mm, top=0.5mm, bottom=0.5mm, left=0mm} 
\definecolor{colorcorefc1}{HTML}{7434EB}
\definecolor{colorcorefc2}{HTML}{9E8B11}
\definecolor{colorcorefc3}{HTML}{119E30}
\newtcbox{\corefmcone}[1][]{enhanced, nobeforeafter, tcbox raise base, boxrule=0.4pt, top=0mm, bottom=0mm, right=0mm, left=3mm, arc=1pt, boxsep=2pt, before upper={\vphantom{dlg}}, colframe=colorcorefc1, coltext=colorcorefc1!75!black, colback=colorcorefc1!10, fonttitle=\bfseries\tiny,#1, overlay={\begin{tcbclipinterior}\fill[colorcorefc1!75] (frame.south west) rectangle node[text=white, font=\sffamily\bfseries\tiny] {1} ([xshift=3mm]frame.north west);\end{tcbclipinterior}}}
\newtcbox{\corefmctwo}[1][]{enhanced, nobeforeafter, tcbox raise base, boxrule=0.4pt, top=0mm, bottom=0mm, right=0mm, left=3mm, arc=1pt, boxsep=2pt, before upper={\vphantom{dlg}}, colframe=colorcorefc2, coltext=colorcorefc2!75!black, colback=colorcorefc2!10, fonttitle=\bfseries\tiny,#1, overlay={\begin{tcbclipinterior}\fill[colorcorefc2!75] (frame.south west) rectangle node[text=white, font=\sffamily\bfseries\tiny] {2} ([xshift=3mm]frame.north west);\end{tcbclipinterior}}}
\newtcbox{\corefmcthree}[1][]{enhanced, nobeforeafter, tcbox raise base, boxrule=0.4pt, top=0mm, bottom=0mm, right=0mm, left=3mm, arc=1pt, boxsep=2pt, before upper={\vphantom{dlg}}, colframe=colorcorefc3, coltext=colorcorefc3!75!black, colback=colorcorefc3!10, fonttitle=\bfseries\tiny,#1, overlay={\begin{tcbclipinterior}\fill[colorcorefc3!75] (frame.south west) rectangle node[text=white, font=\sffamily\bfseries\tiny] {3} ([xshift=3mm]frame.north west);\end{tcbclipinterior}}}

\title{The Role of Natural Language Processing Tasks in Automatic Literary Character Network Construction}

\author{Arthur Amalvy \\
  Laboratoire Informatique d'Avignon \\
  \texttt{arthur.amalvy@univ-avignon.fr} \\\And
  Vincent Labatut \\
  Laboratoire Informatique d'Avignon \\
  \texttt{vincent.labatut@univ-avignon.fr} \\\AND
  Richard Dufour \\
  Laboratoire des Sciences du Numérique de Nantes \\
  \texttt{richard.dufour@univ-nantes.fr}
}

\begin{document}

\maketitle

\begin{abstract}
The automatic extraction of character networks from literary texts is generally carried out using natural language processing (NLP) cascading pipelines. While this approach is widespread, no study exists on the impact of low-level NLP tasks on their performance. In this article, we conduct such a study on a literary dataset, focusing on the role of named entity recognition (NER) and coreference resolution when extracting co-occurrence networks. To highlight the impact of these tasks' performance, we start with gold-standard annotations, progressively add uniformly distributed errors, and observe their impact in terms of character network quality. We demonstrate that NER performance depends on the tested novel and strongly affects character detection. We also show that NER-detected mentions alone miss a lot of character co-occurrences, and that coreference resolution is needed to prevent this. Finally, we present comparison points with 2 methods based on large language models (LLMs), including a fully end-to-end one, and show that these models are outperformed by traditional NLP pipelines in terms of recall.
\end{abstract}

\section{Introduction}
Character networks are graphs whose vertices represent characters, and edges represent the relationships between them. They can be seen as a special case of knowledge graphs, where vertices are restricted to being characters. Such networks have multiple uses: visualize the relationships between characters in a novel, support literary analysis~\citep{rochat-2015-character_networks_zola, rochat-2017, elson-2010-character_networks}, or solve downstream tasks such as recommendation~\citep{lee-2019-charnets} or genre classification~\citep{hettinger-2015-novels_classification}. As indicated by the survey of \citet{labatut-2019}, many authors work on automatically extracting these networks~\citep{sparavigna-2015-chaplin, dekker-2019-character_networks, elson-2010-character_networks}, following a generic three-phase automatic extraction framework. First, identify characters present in a text, using techniques such as NER and alias resolution. Second, detect interactions between characters. One can consider multiple types of interactions (co-occurrence, conversations, actions\ldots), hence this process differs given the targeted type. Third, given characters and their interactions, derive the relationships between characters and extract a character network.

Due to their statistical nature and the difficulty of the tasks they entail, natural language processing (NLP) cascading pipelines applied to character network extraction are bound to make errors. While such networks have been leveraged for many applications, the general question of how to better extract them has been comparatively much less explored, and certainly not in a systemic way. Since extraction is generally performed using an NLP pipeline, the first step to answering this question is to study the impact of each NLP task on the quality of the final network, as understanding it would allow the community to prioritize future research efforts. Therefore, in this article, we propose to study that impact in depth, by artificially adding errors in steps of the extraction pipeline to observe their influence. We specifically focus on NER and coreference resolution, with the latter remaining a challenging task. We perform our study on Litbank~\citep{bamman-2019-litbank,bamman-2020-litbank}, a standard English literary corpus. To support our experiments, we implement extensions to the recent modular character network extraction pipeline Renard~\citep{amalvy-2024-renard}. In order to understand whether cascading pipelines are still competitive against LLM-based extraction systems and to guide future research, we compare our pipeline against such systems. To facilitate reproducibility, we release all our code and data under a free license\footnote{\url{https://github.com/CompNet/Splice}}.

Our contributions are as follows. First, we define new measures to evaluate the quality of extracted networks. Second, we extend the existing Renard extraction pipeline, and evaluate it on a character network dataset we adapt from Litbank. Our measures and dataset are a first step towards a more systematic benchmarking of network extraction systems, something that is missing in the literature. Third, we propose a model to simulate errors for two tasks, NER and coreference resolution, in order to understand their impact on a character network extraction pipeline. Finally, we compare our pipeline to end-to-end LLM models, in order to understand how much cascading errors are detrimental to a sequential pipeline.

We organize the rest of this article as follows. In Section~\ref{sec:related-work}, we discuss related work by highlighting existing character network extraction pipelines and literature on NER and coreference resolution error analysis. In Section~\ref{sec:methods}, we detail our methods, including the extended Renard character network extraction pipeline we use. We describe our experiments in Section~\ref{sec:experiments}, and discuss their results in Section~\ref{sec:results}. Finally, we review our main contributions in Section~\ref{sec:conclusion} and present the limits of our work in Section~\ref{sec:limitations}.

\section{Related Work}
\label{sec:related-work}

\subsection{Character Network Extraction Pipelines}
Character network extraction is related to knowledge graph extraction, but with vertices restricted to characters. Both share common tasks such as entity recognition and linking, but the focus on narratives and characters implies specialized instances with specific challenges.

\textit{BookNLP}~\citep{bamman-2014-booknlp} is a well-known NLP pipeline specialized for novels, and is sometimes used in the literature when it comes to extracting character networks~\citep{dekker-2019-character_networks,piper-2017-character_networks}. Other authors go further and propose pipelines specifically tailored to character network extraction, such as \textit{CHAPLIN}~\citep{sparavigna-2015-chaplin} or \textit{Charnetto}~\citep{metrailler-2023-charnetto}. Recently, \citet{amalvy-2024-renard} propose \textit{Renard}, a modular character network extraction pipeline written in Python. In this article, we extend Renard to conduct a detailed study of each extraction module.

\subsection{Error Analysis}
Most works interested in the effect of NLP errors focus on specific tasks. For the NER task, \citet{stanislawek-2019-ner_glass_ceiling} find that different NER models make different categories of errors, while \citet{rueda-2024-conllsharp} highlight recurrent errors made by models, such as the difficulty of detecting mentions unseen in the training set. For the coreference task, \citet{martschat-2014-coref_error_analysis} focus on recall errors, while \citet{chai-2023-coref_error_analysis} perform an analysis on multilingual coreference systems, and focus on two-mentions entities that they find hard to recall.

To the best of our knowledge, only \citet{dekker-2019-character_networks} adopt a more global view and assess the effect of NER errors on character networks. However, no study exists on the impact of the performance of the main NLP steps required to extract a character network. Our goal in this article is to fill this existing void in the literature by proposing a first impact study.

\section{Methods}
\label{sec:methods}

\subsection{Terminology}

The terminology from the NER, coreference and alias resolution literature diverge and are confusing when used together. This is why, in this section, we clarify the terms that we use in this article. We use \textit{``form''} to refer to a textual representation of a character. A form can be a proper noun (\textit{``Lianna''}), a pronoun (\textit{``she''}), a definite description (\textit{``the princess''})\ldots Meanwhile, we use \textit{``mention''} to refer to the occurrence of a form in the text. A NER system only extracts a subset of characters' mentions: for example, it does not extract pronouns. We refer to the form of a mention detected by a NER model as an \textit{alias}, as it strongly identifies a character. Meanwhile, a coreference system typically detects all mentions, including pronouns and other generic constructs. We therefore distinguish two types of mentions: \textit{alias mentions} and \textit{generic mentions}.

\subsection{Extraction Pipeline}
\label{sec:extraction-pipeline}
To extract character networks, we extend the Renard extraction pipeline~\citep{amalvy-2024-renard}. We design our own pipeline for the needs of this study and contribute different modules. There are many types of interactions that we could extract to produce character networks. As a first study on the subject, we choose to focus on co-occurrence character networks: they are conceptually simple, and are the most used type of networks in the literature~\citep{labatut-2019}. We consider an interaction between two characters when they appear close to each other in the text, in a range we call the \textit{co-occurrence window}. Our pipeline is divided into four main phases: NER, coreference resolution (optional step), character unification, and finally co-occurrence detection and network extraction.

We perform flat NER using the fine-tuned BERT model~\citep{devlin-2019-bert} included in Renard, trained on the literary NER dataset introduced by \citet{dekker-2019-character_networks} and later improved by \citet{amalvy-2023-context_ner}. We only keep mentions of the \texttt{PER} class.

For coreference resolution, we use the end-to-end coreference model included in Renard based on \citet{lee-2017-e2e_coref} and \citet{joshi-2019-bertcoref}. The model predicts links between mentions, but also performs mention detection: this is important when extracting co-occurrence character networks, as generic character mentions (such as pronouns) are still counted as co-occurrences.

Character unification resembles \textit{alias resolution}. In the case of our extraction pipeline, we define character unification as resolving each mention detected by the NER and coreference steps to a single character. This task could be described as a document-level version of coreference resolution, restricted to characters. To unify mentions, we base ourselves on the work of \citet{vala-2015-character_detection}. We construct a graph where each vertex is a character alias as detected by NER, and we employ a set of rules to connect or disconnect these vertices. While rules can introduce errors, they are often used by previous works~\citep{vala-2015-character_detection,ardanuy-2014-novel_clustering}, and thus analyzing their failure modes is important. We use the following rules:
\begin{enumerate}
    \item When two aliases have a first or last name in common, we connect them (\textit{``Emma''} and \textit{``Emma Woodhouse''}).
    \item When two aliases are related by a hypocorism gazetteer (\textit{``John''} and \textit{``Johnny''}), we connect them.
    \item When one of the two above rules holds for two aliases when removing titles, we connect them (\textit{``Mr. John''} and \textit{``Johnny''}).
    \item When two aliases are coreferential, we connect them. We consider two aliases to be coreferential when they appear together in one or more coreference chains, and never appear without the other in other chains.
    \item When connected aliases have the same last name but a different first name, we delete all vertices in the shortest paths between them, since they are probably different characters from the same family (\textit{``John Smith''} and \textit{``John Klint''}).
    \item When two aliases have a different inferred gender, we delete all the edges in the shortest paths between them (\textit{``Mr. Smith''} and \textit{``Miss Smith''}). We infer gender using the gendered titles and pronouns in coreference chains.
\end{enumerate}
After having applied all these rules, we merge the graph-connected components. Using the alias groups extracted with this algorithm, we assign each mention detected by the NER or coreference steps to a single character.

Finally, we apply the co-occurrence detection and network extraction step of Renard. This step is entirely deterministic and cannot cause any errors errors by itself: we simply consider that two character mentions in the defined co-occurrence window form an interaction, which results in an edge between these characters. To take into account the importance of each relationship, we weight edges by the number of interactions between characters.

\subsection{Perturbation Analysis}
\label{sec:methods-perturbation-analysis}
To assess the impact of NER and coreference resolution errors on the extracted networks, we propose to start from a pipeline with gold-standard NER and coreference predictions, and to progressively degrade the performance of these tasks while observing the impact on the quality of the extracted networks. To degrade task performance, we add uniformly distributed perturbations to the predictions, corresponding to different types of errors.

\subsubsection{NER Perturbations}

As an example, we consider the following gold predictions as a starting point, and consider two types of perturbations.
\begin{exampleNER}
  \tagPER{One-Eye} \tagO{looked} \tagO{at} \tagPER{Goblin} \tagO{.}
\end{exampleNER}

\paragraph{Add Spurious Alias Mentions:} We add false positives to the NER predictions by uniformly sampling generic spans (up to a certain span size) from the text, to reduce Precision.
\begin{exampleNER}
  \tagPER{One-Eye} \tagPER{looked} \tagO{at} \tagPER{Goblin} \tagO{.}
\end{exampleNER}

\paragraph{Remove Correct Alias Mentions:} We remove true positives from the NER predictions by uniformly sampling from the predicted alias mentions, to reduce Recall.
\begin{exampleNER}
  \tagPER{One-Eye} \tagO{looked} \tagO{at} \tagO{Goblin} \tagO{.}
\end{exampleNER}

\subsubsection{Coreference Resolution Perturbations}
\label{sec:methods-coref-perturbation}

As an example, we consider the following gold prediction as a starting point, and consider four types of perturbations.
\begin{examplecoref}
  \corefmcone{One-Eye} pranced over and took a poke at \corefmctwo{Goblin}, trying to break \corefmctwo{his} concentration.
\end{examplecoref}

\paragraph{Add Spurious Mentions:} We add singletons (mentions linked to no other mentions) to the predictions consisting of incorrect mentions, by uniformly sampling non-mention spans (up to a certain span size).
\begin{examplecoref}
  \corefmcone{One-Eye} pranced over and took a \corefmcthree{poke} at \corefmctwo{Goblin}, trying to break \corefmctwo{his} concentration.
\end{examplecoref}

\paragraph{Remove Correct Mentions:} We remove correctly predicted mentions from the predictions by uniform sampling.
\begin{examplecoref}
  One-Eye pranced over and took a poke at \corefmctwo{Goblin}, trying to break \corefmctwo{his} concentration.
\end{examplecoref}

\paragraph{Add Spurious Links:} We add incorrect coreference links between two mentions, wrongly merging coreference chains together. We uniformly sample the incorrect links in the set of all possible incorrect links.
\begin{examplecoref}
  \corefmctwo{One-Eye} pranced over and took a poke at \corefmctwo{Goblin}, trying to break \corefmctwo{his} concentration.
\end{examplecoref}

\paragraph{Remove Correct Links:} We remove correct links between predicted mentions, wrongly splitting coreference chains. We uniformly sample links among all existing correct coreference links.
\begin{examplecoref}
  \corefmcone{One-Eye} pranced over and took a poke at \corefmctwo{Goblin}, trying to break \corefmcthree{his} concentration.
\end{examplecoref}

\subsection{Network Quality Measures}
\label{sec:graph-quality-measures}
Since we want to measure the impact of NLP errors on the extracted network, we need a set of measures to assess the quality of this network when compared to a reference network. We base our measures on the work of~\citet{vala-2015-character_detection} on alias resolution.

Let $G_p = (V_p, E_p)$ be a predicted character network, and $G_g = (V_g, E_g)$ be the corresponding gold network. Let each vertex of $V_p$ and $V_g$ represent the set of aliases $\{a_1, a_2, \cdots, a_n\}$ of the underlying character. In order to know whether the predicted network $G_p$ correctly contains vertices and edges similar to the gold network $G_g$, we first need to match their characters, since a vertex in $V_p$ is not necessarily present in $V_g$ and vice versa. Thus, we start by computing a maximum bipartite mapping $f_V$ from the set of predicted vertices $V_p$ to $V_g \cup \{ v_{\varnothing} \}$. This mapping associates any predicted vertex $u \in V_p$ to a gold vertex $v \in V_g$ or to the null vertex $v_{\varnothing}$, meaning $u$ is not associated with any character in $V_g$. Note that the null vertex $v_{\varnothing}$ represents the empty set of aliases. Symmetrically, we construct a mapping $g_V$ from $V_g$ to $V_p \cup \{ v_{\varnothing} \}$. We leverage the alias sets represented by the vertices to compute Vertex Precision and Vertex Recall\footnote{We employ the Iverson bracket notation, where $[P] = 1$ if proposition $P$ is true, and $0$ otherwise.}:

\begin{align}
    Pre_V = \max_{f_V} \frac{\sum_{u \in V_p}{1 - \frac{|u - f_V(u)|}{|u|}}}{|V_p|} \\
    Rec_V = \max_{g_V} \frac{\sum_{v \in V_g}{[g_V(v) \cap v \neq v_{\varnothing}]}}{|V_g|}.
\end{align}

We define Vertex F1 ($F1_V$) as the harmonic mean between Vertex Precision and Vertex Recall.

For edges, we use mappings $f_V$ and $g_V$ to construct mappings $f_E$ and $g_E$, which map similarly edge sets $E_p$ and $E_g$:
\begin{equation}
  f_E  (\{u, v\}) = 
  \begin{cases}
    \{f_V(u), f_V(v)\}, & \text{if } f_V(u) \neq v_{\varnothing}  \\
                        & \text{and } f_V(v) \neq v_{\varnothing} \\
    v_{\varnothing},    & \text{otherwise.}
    \nonumber
  \end{cases}
\end{equation}
Based on $f_E$ and $g_E$, we compute Edge Precision and Edge Recall:
\begin{align}
    Pre_E = \max_{f_E} \frac{\big|\{f_E(e) : e \in E_p\} \cap E_g\big|}{|E_p|} \\
    Rec_E = \max_{g_E} \frac{\big|E_p \cap \{g_E(e) : e \in E_g\}\big|}{|E_g|}.
\end{align}
We define Edge F1 ($F1_E$) as the harmonic mean of Edge Precision and Edge Recall.

We also introduce weighted variants of these network measures ($WPre_E$, $WRec_E$ and $WF1_E$) in order to take into account the weights of the network edges, that correspond to the number of interactions between connected characters. Before computing the measures, we normalize the weights by dividing by the maximal number of co-occurrences in the network. We compute Precision and Recall as follows:

{\small
\begin{align}
    WPre_E = \max_{f_E} \frac{\sum_{e \in E_p}{1 - \big|w(f_E(e)) - w(e)\big|}}{|E_p|} \\
    WRec_E = \max_{g_E} \frac{\sum_{e \in E_g}{1 - \big|w(e) - w(g_E(e))\big|}}{|E_g|},
\end{align}
}

where $w(e)$ is the function that computes the normalized weight of edge $e$. $w(e)$ is $0$ when $e = e_{\varnothing}$. Weighted measures evaluate the quality of the distribution of weights in the predicted network, and are always less than or equal to their unweighted counterparts.

\section{Experiments}
\label{sec:experiments}

\subsection{Literary Corpus}
\label{sec:corpus}
We perform all of our experiments on the NER and coreference layers of the Litbank literary corpus~\citep{bamman-2019-litbank,bamman-2020-litbank}. Since it is designed for nested NER while the Renard NER step performs flat NER, we flatten the Litbank annotations using an algorithm we implement (see Appendix~\ref{apdx:adapting-litbank} for details). We use coreference chains of \texttt{PER} mentions as the ground truth for the character unification step, since coreference resolution on characters is equivalent to character unification. As the network extraction step cannot cause errors on its own, we extract gold character networks using these annotations only.

Litbank is composed of excerpts from 100 novels of approximately 2,000 tokens each. Since these excerpts can be short, we restrict our analysis to the 30 novels involving at least 10 characters. This prevents high deviation of network quality measures when a vertex or an edge is modified in the prediction. We use the remaining 70 novel excerpts to train coreference resolution models: we use 63 of these 70 novels (90\%) as a training set, and the remaining 7 as a development set.

\subsection{Pipeline Performance}
We apply our character network extraction pipeline to the 30 excerpts we select for analysis. Since multiple coreference resolution measures exist and none of these are entirely satisfying or measure the same thing, we report a large set of measures including MUC~\citep{vilain-1995-coref_muc}, $B^3$~\citep{bagga-1998-b3}, CEAF~\citep{luo-2005-ceaf}, BLANC~\citep{recasens-2011-blanc} and LEA~\citep{moosavi-2016-lea}. We also report the performance of a pipeline without the optional coreference resolution step, in order to assess the usefulness of the task. We consider a co-occurrence window of 32 tokens.

\subsection{LLM-based Approaches}
\label{sec:llm-pipelines}

Since character network extraction is usually performed using cascading pipelines, errors may propagate and degrade performance. Meanwhile, LLM-based approaches are less modular and explainable, but are not subject to cascading errors by design. Therefore, inspired by the recent advances in LLMs, we survey their capability to be used as character network extractors, and compare them to our cascading Renard pipeline. We introduce two different LLM extraction methods:

\paragraph{LLM-Coref} We remark that the last network extraction step, co-occurrence detection, cannot cause errors on its own: we simply create an edge between two characters if some of their mentions are in the same co-occurrence window. Therefore, in the case of a co-occurrence network, we can define the extraction problem as span extraction, where each extracted span must be assigned to the correct character. This problem definition could also be viewed as coreference resolution restricted to characters only. To tackle the problem this way, we prompt LLMs to mark character mentions in the text with a unique character ID.

\paragraph{LLM-E2E} Given an input text, we simply prompt LLMs to produce the corresponding character network in a simplified version of the XML-based Graphml format.

See Appendix~\ref{apdx:llms-prompt} for the exact prompts. We use few-shot prompting by providing examples of the task to the model. For both methods, we make an effort to parse the LLM output in order to fix slightly incorrect output format. We survey two proprietary models, GPT3.5 Turbo~\citep{brown-2020-gpt3}\footnote{GPT3.5 checkpoint: \texttt{gpt-3.5-turbo-0125}} and GPT4o\footnote{GPT4o checkpoint: \texttt{gpt-4o-2024-05-13}}, and a recent open weights model, Llama3-8b-instruct~\citep{touvron-2023-llama}.

\section{Results}
\label{sec:results}

\subsection{Pipeline Performance}
Table~\ref{tab:perf_tasks} shows the NER and coreference resolution performance of our extended Renard pipeline. NER performance is below the reported state-of-the-art on datasets from other domains such as CoNLL-2003~\citep{tjong-2003-conll_2003_ner} where the best systems obtain F1 scores higher than 90. While part of this lack of performance may be due to the way we transform the nested Litbank dataset to a flat NER dataset, we observe a high disparity of performance between novels, with F1 ranging from $51.16$ to $97.30$. This matches the observations from \citet{dekker-2019-character_networks} and \citet{amalvy-2023-learning_to_rank}, indicating challenges that are specific to some novels. Meanwhile, the performance of coreference resolution is lower than what \citet{bamman-2020-litbank} reports on Litbank (for example, we report $78.45$ MUC while \citet{bamman-2020-litbank} reports $84.3$). This may be due to the lower number of excerpts in our training set (63 vs. 80), which is required for our analysis.

\begin{table}[tbh]
  \centering
  \begin{tabular}{ll@{\hskip 1cm}rrr}
    \hline
    \textbf{Task}  & \textbf{Measure} & \textbf{Mean}    & \textbf{Min}     & \textbf{Max} \\ 
    \hline
    NER            & F1               & $79.58$          & $51.16$          & $97.30$      \\
    \hline 
    Coref          & MUC              & $78.45$          & $64.31$          & $88.75$      \\
                   & B3               & $54.87$          & $41.53$          & $70.40$      \\
                   & CEAF             & $47.04$          & $34.31$          & $59.75$      \\
                   & BLANC            & $60.82$          & $40.64$          & $79.82$      \\
                   & LEA              & $28.84$          & $19.44$          & $43.95$      \\
    \hline
  \end{tabular}
  \caption{Performance of our pipeline on NER and coreference resolution. We compute the Mean, Min and Max values on the series formed by the measures of the 30 novel excerpts of our analysis set.}
  \label{tab:perf_tasks}
\end{table}

Table~\ref{tab:perf_coref_nocoref} shows the performance of our pipeline on our test corpus of 30 excerpts, depending on whether we add the optional coreference step or not. Vertex and edge F1 are higher when omitting coreference information, likely because the performance of the coreference resolution algorithm is not high enough, leading to the detection of spurious mentions, which misleads both the character unification and co-occurrence detection steps. We discuss the question of the utility of coreference resolution in more detail in Section~\ref{sec:coref-perf}. In general, Edge Recall is quite low, meaning many character interactions are missed.

\begin{table}[tbh]
  \centering
  \begin{tabular}{l@{\hskip 1cm}rr}
    \hline
    \textbf{Measure} & \textbf{w/ coref} & \textbf{w/o coref} \\
    \hline
    $F1_V$           & $57.64$           & $\mathbf{70.39}$   \\
    $F1_E$           & $40.19$           & $\mathbf{44.93}$   \\
    $WF1_E$          & $\mathbf{33.53}$  & $30.55$            \\
    \hline 
    $Pre_V$          & $59.32$           & $\mathbf{68.99}$   \\
    $Pre_E$          & $48.07$           & $\mathbf{62.07}$   \\
    $WPre_E$         & $38.40$           & $\mathbf{39.91}$   \\
    \hline 
    $Rec_V$          & $57.77$           & $\mathbf{74.00}$   \\
    $Rec_E$          & $\mathbf{39.37}$  & $37.81$            \\
    $WRec_E$         & $\mathbf{33.17}$  & $26.18$            \\
    \hline
  \end{tabular}
  \caption{performance of our pipeline on network extraction with or without the coreference resolution step.}
  \label{tab:perf_coref_nocoref}
\end{table}

Meanwhile, using coreference information proves to be important to increase edge recall measures. Even though coreference resolution leads to a compromised network structure overall, it allows the pipeline to detect more character mentions, leading to a better estimation of the relative strength of their interactions.

\subsubsection{Coreference Resolution Performance}
\label{sec:coref-perf}
Given the negative impact that coreference resolution can have, as seen in Table~\ref{tab:perf_coref_nocoref}, a question that naturally arises is whether this task is useful when extracting character networks. Co-occurrence networks extracted with alias mentions might only be a sufficiently good approximation of a network extracted with all mentions. In that case, performing a coreference resolution to extract additional mentions would not be critical.

\begin{table}[tbh]
  \centering
  \begin{tabular}{l@{\hskip 1cm}rrr}
    \hline
    \textbf{Measure} & $Rec_E$ & $WPre_E$ & $WRec_E$ \\
    \hline
    \textbf{Value}   & $54.39$ & $63.20$  & $35.66$  \\
    \hline
  \end{tabular}
  \caption{Network quality measures for gold networks extracted by ignoring coreference mentions. Only affected measures are presented.}
  \label{tab:gold_coref_vs_nocoref}
\end{table}

To understand whether this is the case, we extract gold networks from Litbank with and without coreference-extracted mentions, and compute network quality measures by considering networks with coreference mentions as the reference. Results can be found in Table~\ref{tab:gold_coref_vs_nocoref}. While only some measures are affected (Edge Recall, Weighted Edge Precision, and Weighted Edge Recall), ignoring coreference mentions proves to severely impact performance. This shows that coreference mentions are a crucial part of relationships extracted using co-occurrence.

\begin{figure}[tbh]
  \centering
  \includegraphics[width=\linewidth]{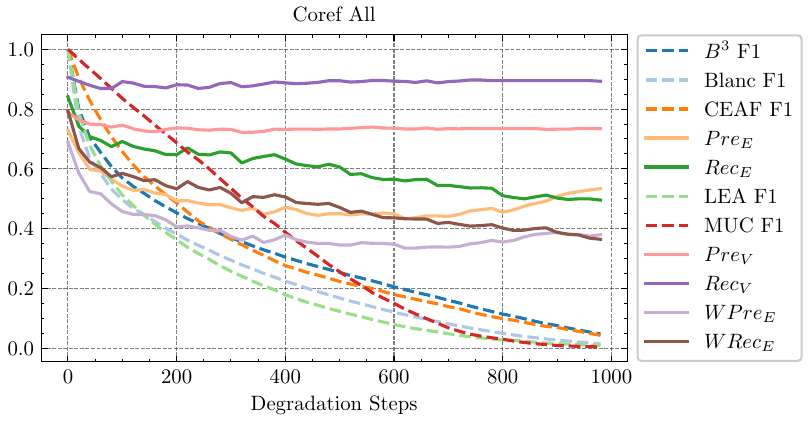}
  \caption{Network extraction performance when applying the coreference perturbations from Section~\ref{sec:methods-coref-perturbation}.}
  \label{fig:coref_all}
\end{figure}

To try and understand the minimal performance needed for coreference resolution to be useful, we perform an experiment where we inject the gold coreference information in the pipeline, and slowly degrade performance. To do so, we combine all the coreference perturbations we described in Section~\ref{sec:methods-coref-perturbation}, by uniformly sampling a degradation and applying it. We repeat this process for a fixed number of steps and observe the effects in terms of coreference and network quality measures. The results of this experiment can be found in Figure~\ref{fig:coref_all}.

When applying perturbations, overall performance for all measures starts to decrease until a certain point, after which it starts reaching a plateau (except for edge recall measures). This effect occurs because after this point, coreference resolution recall on alias mentions starts being so low that it does not affect the character unification algorithm anymore, and other rules based on character aliases start to have more influence. Meanwhile, there is a big discrepancy between vertex and edge measures: while vertex measures can stay close to their original values when coreference resolution is highly degraded, this is not the case for edge measures. This is because coreference resolution is crucial when detecting co-occurrence, as it is the only way to detect generic mentions.

\subsubsection{NER Perturbation Analysis}
\label{sec:perturbation-analysis-tasks}

\begin{figure*}[tbh]
  \centering
  \includegraphics[width=0.9\textwidth]{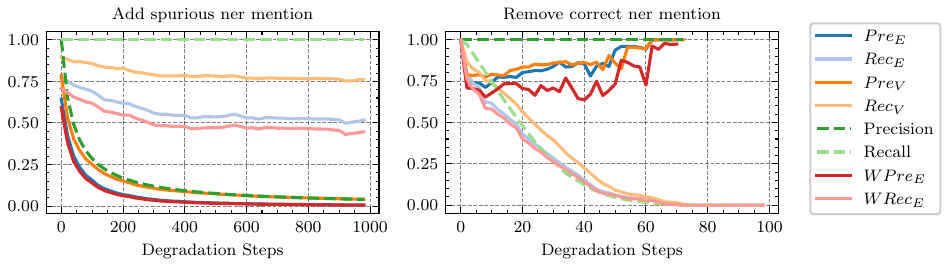}
  \caption{Network quality measures versus number of degradation steps for ``add spurious alias mention'' and ``remove correct alias mention'' perturbations.}
  \label{fig:ner-degradation}
\end{figure*}

Figure~\ref{fig:ner-degradation} shows the NER perturbation results.

\paragraph{Add Spurious Alias Mentions} Unsurprisingly, reducing NER Precision has a direct effect on Vertex Precision, which plummets as more NER false positives are added. Edge Precision also sharply decreases as a result, while Vertex and Edge Recall slowly decrease down to a plateau.

\paragraph{Remove Correct Alias Mentions} All recall measures sharply decrease when removing correct alias mentions. Precision measures become unstable and finally undefined, as no vertices or edges are predicted when NER Recall reaches $0$.

\medskip
Unsurprisingly, NER performance has a high impact on network quality. Since NER performance varies greatly depending on the novels (as seen in Table~\ref{tab:perf_tasks}), enhancing performance for challenging novels is an important concern that should be addressed by future research.

\subsubsection{Coreference Resolution Perturbation Analysis}
\begin{figure*}[tbh]
  \centering
  \includegraphics[width=1\textwidth]{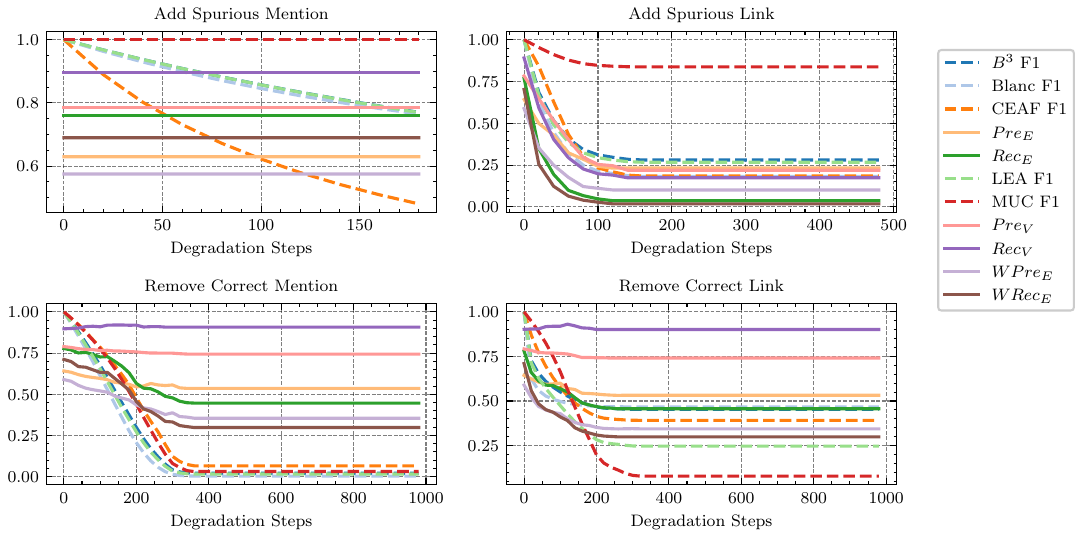}
  \caption{Network quality measures versus the number of coreference resolution degradation steps.}
  \label{fig:coref-degradation}
\end{figure*}

Figure~\ref{fig:coref-degradation} shows the coreference resolution perturbations results. 

\paragraph{Add Spurious Mentions} Adding spurious singletons does not affect our character unification algorithm, and therefore has no impact on the quality of the extracted network.

\paragraph{Remove Correct Mentions} Removing correct mentions mainly affects edge measures: characters are still recognized correctly, but some co-occurrence interactions are lost due to missing character mentions, leading to fewer edges.

\paragraph{Add Spurious Links} Adding wrong coreference links sharply decreases all network extraction performance measures: characters are harder to recognize, and interactions are missing. This is driven by rules 4 and 6 of our character unification step (Section~\ref{sec:extraction-pipeline}), that are both fed wrong information. While it would be possible to not apply these rules, rule 4 is the only one that allows linking two mentions with completely different forms.

\paragraph{Remove Correct Links} Network edge measures are affected the most by coreference links removal, while vertex measures stay somewhat stable.

\medskip
Not all coreference resolution errors prove to be equal in terms of impact on network quality. Adding spurious links is the most harmful error, while adding spurious singletons does not affect our character unification algorithm. Meanwhile, other errors mainly impact edge extraction performance. Therefore, if one's concern is to extract characters only, a conservative coreference algorithm with low linking recall, but high linking precision might give sufficient performance. However, when extracting edges between characters, both high precision and recall are necessary, in terms of mention detection as well as linking.

\subsection{LLM-based Approaches}

\begin{table*}[tbh]
\centering
\begin{tabular}{l@{\hskip 1cm}rrr@{\hskip 1cm}rrr@{\hskip 1cm}r}
\hline
                & \multicolumn{3}{c}{\textit{LLM-Coref}} & \multicolumn{3}{c}{\textit{LLM-E2E}} & \textit{Renard}                                                                                             \\
\textbf{Measure} & \textbf{Llama3}                        & \textbf{GPT-3.5T}                        & \textbf{GPT4o}   & \textbf{Llama3} & \textbf{GPT-3.5T} & \textbf{GPT4o}     &                  \\ 
\hline      
$F1_V$          & $37.93$                                & $28.99$                                  & $52.32$          & $56.87$         & $44.26$           & $62.98$            & $\mathbf{70.39}$ \\
$F1_E$          & $23.20$                                & $16.96$                                  & $38.85$          & $29.35$         & $20.91$           & $30.42$            & $\mathbf{44.93}$ \\
$WF1_E$         & $15.17$                                & $11.39$                                  & $\mathbf{32.72}$ & $20.17$         & $14.33$           & $24.35$            & $30.55$          \\
\hline
$Pre_V$         & $42.86$                                & $52.50$                                  & $68.78$          & $67.96$         & $61.04$           & $\mathbf{77.96}$   & $68.99$          \\
$Pre_E$         & $57.85$                                & $\mathbf{65.78}$                         & $62.62$          & $59.01$         & $64.39$           & $65.76$            & $62.07$          \\
$WPre_E$        & $34.69$                                & $37.25$                                  & $51.46$          & $40.56$         & $44.75$           & $\mathbf{53.97}$   & $39.91$          \\
\hline
$Rec_V$         & $25.12$                                & $22.15$                                  & $32.28$          & $53.18$         & $37.87$           & $34.24$            & $\mathbf{70.39}$ \\
$Rec_E$         & $10.34$                                & $9.18 $                                  & $23.38$          & $21.32$         & $13.24$           & $13.12$            & $\mathbf{37.81}$ \\
$WRec_E$        & $6.86$                                 & $6.34$                                   & $19.77$          & $14.86$         & $ 9.14$           & $10.44$            & $\mathbf{26.18}$ \\
\hline
\end{tabular}
\caption{Comparison between the network extraction performance of LLM-based extraction methods and our Renard pipeline. \textit{Llama3} stands for Llama3-8b-instruct, \textit{GPT-3.5T} for GPT-3.5 Turbo.}
\label{tab:perf_e2ellm}
\end{table*}

Results of our LLM-based extraction methods \textit{LLM-Coref} and \textit{LLM-E2E} can be found in Table~\ref{tab:perf_e2ellm}.

\paragraph{LLM-Coref} If we observe F1 scores, GPT4o performs the best amongst the LLMs we survey, followed by Llama3-8b-instruct and GPT-3.5 Turbo. LLMs particularly struggle with recall, with GPT3.5 Turbo and Llama3-8b-instruct missing a lot of character occurrences. Qualitatively, both of these models either miss a lot of mentions, or start hallucinating a lot of mentions before continuing that way through the output, influenced by their previous predictions. Llama3-8b-instruct also has trouble respecting the output format, sometimes generating invalid output. GPT4o is much more consistent, although it still misses many mentions. When compared against our extended Renard pipeline, LLMs results are generally lower, except for Edge Precision and for weighted edge measures where GPT4o trades precision for recall. 

\paragraph{LLM-E2E} Results are generally higher than with \textit{LLM-Coref}, highlighting the importance of task formulation. If we focus on F1 scores, we observe the same ranking between LLMs, with GPT4o beating Llama3-8b-instruct and GPT3.5 Turbo. Generally, LLMs display high vertex and edge precision, even surpassing our pipeline sometimes. However, their recall still lags behind Renard.

\section{Conclusion}
\label{sec:conclusion}
In this article, we presented a study on the impact of NLP tasks on character network extraction. We show that NER performance is crucial to detecting characters, but that it depends heavily on the considered novel: therefore, future research should focus on improving the low NER performance on difficult novels. Additionally, we show that not tackling the challenging coreference resolution task implies missing co-occurrence relationships between characters. This task is important to extract correct co-occurrence edges, particularly when it comes to edge weights. Since it remains difficult in general, our extraction pipeline exhibits relatively low edge extraction performance. We also show that not all coreference errors have the same impact: adding spurious coreference links between mentions has the strongest negative impact of all the errors we surveyed. Developing systems able to make conservative predictions when it comes to coreference linking might be a good research direction to create better character network extraction systems. Unfortunately, developing coreference models at the scale of a novel remains difficult due to the lack of fully annotated ones for training and benchmarking, which prompts the development of datasets with long documents.

Even though errors propagate in cascading pipelines, our pipeline generally outperformed LLM-based approaches. However, the performance of these approaches is encouraging given the fact that we only evaluated the few-shot prompting setting. Fine-tuning LLMs on character network extraction is therefore a promising direction of research, even though pipelines remain more interpretable.

\FloatBarrier

\section{Limitations}
\label{sec:limitations}
\begin{itemize}
  \item While the character network extraction pipeline we use is inspired by the generic framework outlined by \citet{labatut-2019}, we still had to make implementation choices. Other pipelines may behave differently regarding task errors, although we hypothesize that similar architectures should yield similar results.
  \item Our perturbation analysis methodology may not reflect the distribution of errors from existing models. However, it allows considering the different types of possible errors.
  \item For end-to-end extraction using LLMs, we only survey the few-shot prompting setting due to resource limitations. Fine-tuning a model may yield better results. However, the excerpts on which we perform analysis are short (approximately 2,000 tokens) compared to full-scale novels: the performance of LLMs in that setting may not be as high as the results we report in our study.
\end{itemize}

\bibliography{Amalvy2024.bib}

\appendix

\section{Large Language Models Prompt}
\label{apdx:llms-prompt}

We use the following prompt for our \textit{LLM-Coref} and \textit{LLM-E2E} extraction methods respectively:

\begin{promptenv}
\textbf{SYSTEM PROMPT} You are an expert in literature and natural language processing.

\textbf{USER PROMPT} Given a text, you must extract characters and their mentions. Your answer must be the original text, where character mentions are tagged with the following format: [CHARACTER\_ID]CHARACTER MENTION[/CHARACTER\_ID]. You must tag character mentions only.

\vspace{0.5cm}
Here are some examples of this task:

Example 1:

Input:
Elric was riding his horse . Alongside Moonglum , the prince of ruins was looking for his dark sword .

Output:
[0] Elric [/0] was riding [0] his [/0] horse . Alongside [1] Moonglum [/1] , the [0] prince of ruins [/0] was looking for [0] his [/0] dark sword .

\vspace{0.5cm}
Example 2:

Input:
Princess Liana felt sad , because Zarth Arn was gone . The princess decided she should sleep .

Output:
[0] Princess Liana [/0] felt sad , because [1] Zarth Arn [/1] was gone . [0] The princess [/0] decided [0] she [/0] should sleep .
\end{promptenv}

\begin{promptenv}
    
\textbf{SYSTEM PROMPT} You are an expert in literature and natural language processing.

\textbf{USER PROMPT} 
Given a text, you must extract a co-occurrence character network where vertices represent characters and edges represent their relationships. Each edge must have a weight corresponding to the number of interactions between two characters. Two characters without any interactions do not share an edge. An interaction between two characters occurs when two of their mentions occur within a distance of 32 tokens.

Your answer must be in a simplified Graphml-like format. Vertices must have an 'alias' attribute with the list of aliases of a character, separated by semicolons.

\vspace{0.5cm}
Here are some examples of this task:

Example 1:

Input:
Elric was riding his horse . Alongside Moonglum , the prince of ruins was looking for his dark sword .

Output:
\begin{lstlisting}
<graph>
<node id="n0"
 aliases="Elric;prince of ruins">
</node>
<node id="n1" aliases="Moonglum">
</node>
<edge id="e0" source="n0"
 target="n1"
 weight="2">
</edge>
</graph>
\end{lstlisting}

\vspace{0.5cm}
Example 2:

Input:
Princess Liana felt sad , because Zarth Arn was gone . Liana decided she should sleep .

Output:
\begin{lstlisting}
<graph>
<node id="n0"
 aliases="Princess Liana;Liana">
</node>
<node id="n1" aliases="Zarth Arn">
</node>
<edge id="e0" source="n0" target="n1"
 weight="2">
</node>
</graph>
\end{lstlisting}

\end{promptenv}

\section{Adapting Litbank to Flat NER}
\label{apdx:adapting-litbank}

In the main body of this article, we perform our experiments on the Litbank dataset~\citep{bamman-2019-litbank}. However, Litbank NER annotations are nested, while the Renard NER step we employ performs flat NER. Annotations guidelines are also different between Litbank and the original NER dataset on which the Renard NER step is trained~\citep{amalvy-2023-context_ner}. In particular, Litbank annotates many generic mentions (such as \textit{``an honourable man''}) as alias mentions. We therefore flatten Litbank annotations using an algorithm we implement, while trying to respect the original annotation guidelines as much as possible.

Litbank annotations consists of 4 layers of nesting, where annotated alias mentions can overlap. Our flattening algorithm works as follows: first, we try to cut annotated mentions to a form that would be accepted as an alias mention in the dataset of \citet{amalvy-2023-context_ner}. We do so by removing leading determiners (\textit{``the''}) and cutting the content of a mention after the first comma. Afterward, we filter mentions that are still deemed generic, by checking if their constituting tokens are capitalized (except for some stopwords). If some alias mentions are still overlapping at this point, we select the outermost ones.

To give an example, the annotated mention \textit{``the Lord High Chancellor''} would have been shortened as \textit{``Lord High Chancellor''}, and then accepted as an alias mention since all of its tokens are capitalized (provided it does not overlap with a larger mention). By contrast, the generic mention \textit{``an honourable man''} would have been discarded since its tokens are not capitalized.

\end{document}